\documentclass[10pt,twocolumn,letterpaper]{article}

\usepackage{cvpr}              
\usepackage{amssymb}
\usepackage{pifont}
\usepackage{dsfont}
\newcommand{\cmark}{\ding{51}}%
\newcommand{\xmark}{\ding{55}}%
\usepackage[bb=boondox]{mathalfa}
\usepackage{bbold}
\usepackage[dvipsnames]{xcolor}

\definecolor{cvprblue}{rgb}{0.21,0.49,0.74}
\usepackage[pagebackref,breaklinks,colorlinks,citecolor=cvprblue]{hyperref}
\usepackage{multirow}
\def\paperID{11} 
\def\confName{3DMV}
\def\confYear{2024}

\title{\textit{Cross-Mo}dal \textit{S}elf-\textit{T}raining: Aligning Images and Pointclouds to learn Classification without
Labels}

\author{Amaya Dharmasiri\\
Princeton University\\
\and
Muzammal Naseer\\
MBZUAI\\
\and
Salman Khan\\
MBZUAI\\
\and
Fahad Shahbaz Khan\\
MBZUAI\\
}

\begin{document}
\maketitle

\begin{abstract}
Large-scale vision 2D vision language models, such as CLIP can be aligned with a 3D encoder to learn generalizable (open-vocabulary) 3D vision models. However, current methods require supervised pre-training for such alignment, and the performance of such 3D zero-shot models remains sub-optimal for real-world adaptation. In this work, we propose an optimization framework: \emph{\textbf{Cross-MoST: }\textbf{Cross-Mo}dal \textbf{S}elf-\textbf{T}raining}, to improve the label-free classification performance of a zero-shot 3D vision model by simply leveraging unlabeled 3D data and their accompanying 2D views. We propose a student-teacher framework to simultaneously process 2D views and 3D point clouds and generate joint pseudo labels to train a classifier and guide cross-model feature alignment. Thereby we demonstrate that 2D vision language models such as CLIP can be used to complement 3D representation learning to improve classification performance without the need for expensive class annotations. Using synthetic and real-world 3D datasets, we further demonstrate that \textbf{Cross-MoST} enables efficient cross-modal knowledge exchange resulting in both image and point cloud modalities learning from each other's rich representations. The code and pre-trained models are available \href{https://github.com/theamaya/CrossMoST}{here}.
\vspace{-10pt}
\end{abstract}

\section{Introduction}
\label{sec:introduction}

Recent developments of foundational models such as CLIP \cite{CLIP}, contrastively pre-trained on large-scale 2D image text pairs, enable open-vocabulary zero-shot classification. On the other hand, the 3D visual domain which has an increasingly important role in real-world applications such as mixed reality, robotics, and autonomous driving, suffers from the limited amount of training data. 
Therefore, directly learning 3D foundational models lacks scalability. Recent works \cite{cg3d, ULIP} propose to train a 3D encoder by aligning its latent space with a pre-trained 2D image-text model, CLIP \cite{CLIP}. 
This allows zero-shot 3D classification whose performance, however, remains sub-optimal for real-world adaptation, especially when compared to supervised learning. 
Nevertheless, supervised learning requires expensive annotated datasets.\cite{shapenet, deitke2022objaverse}.

\begin{figure}[!t]
\centering
    \includegraphics[width=1\columnwidth]{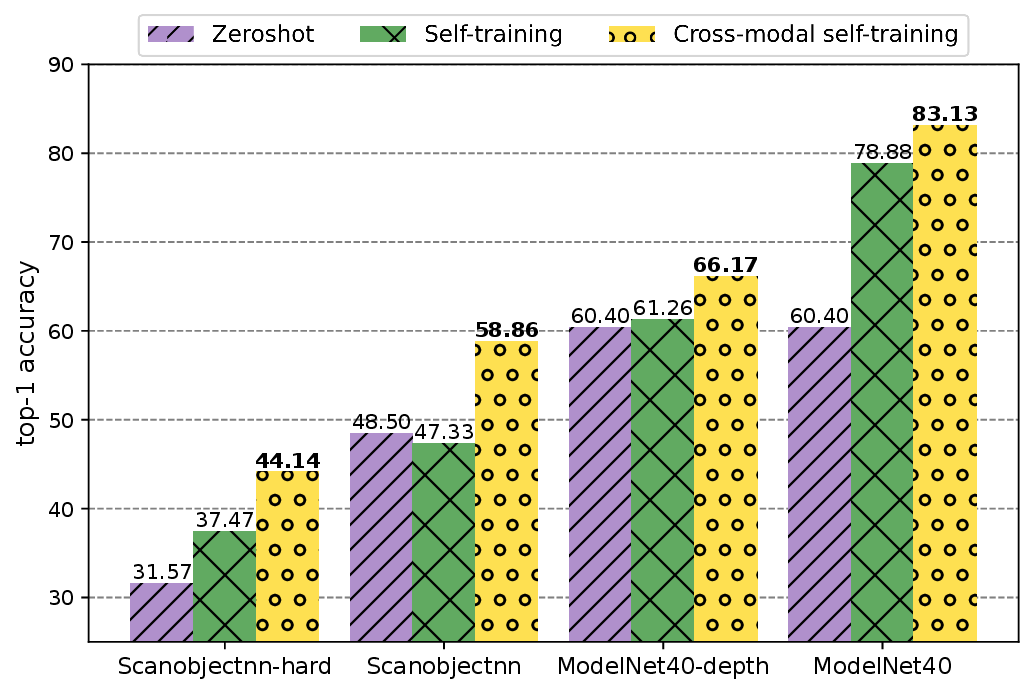}
    \caption{\small \textbf{Proposed cross-modal self-training} achieves significant performance gains over zero-shot \cite{ULIP} 3D classification, as well as recently proposed self-training\cite{MUST} applied on point clouds.}
  \label{fig:concept_figure_bargraph}
\vspace{-10pt}
\end{figure}

Self-training is an interesting learning paradigm that belongs to semi-supervised learning and aims to adapt models for downstream tasks where pseudo-labels from unlabeled data are used as training targets. 
Especially for large foundational models such as CLIP pretrained on very general datasets, self-training acts as a useful training paradigm to adopt its general knowledge to specific downstream tasks without requiring any labels. 
Self-training on images capitalizing on the open-vocabulary zero-shot classification ability of CLIP to generate a pseudo-supervisory signal has been explored by works such as MUST\cite{MUST}.
However, adopting similar self-training methods in other modalities such as 3D point clouds has not been widely explored, and is accompanied by the challenge of noise in pseudo labels due to the lack of pretrained knowledge and limited open-vocabulary performance.

On the other hand, real-world data gathered by 3D scanners are often accompanied by their corresponding RGB and/or RGBD images, whereas synthetic 3D data such as CAD models can easily be rendered into a set of defined 2D views. This provides an opportunity for two coexisting data modalities to learn from each other's unique representations to understand one reality; even without labels.

To this end, we propose  "\textit{Cross-MoST: Cross-Modal Self-Training:} Aligning Images and Pointclouds to learn Classification without Labels" aiming to 1) Harness the multimodality of data to mitigate the lack of expensive annotations, 2) Generate more robust pseudo-labels to enable self-training on 3D point clouds, and 3) Implement cross-modal learning and facilitate both image and point cloud modalities to learn from each other's unique and rich representations. 
Our contributions are as follows:

\begin{itemize}
\item We explore self-training as a setting to implement cross-modal learning. 
By formulating joint pseudo-labels by taking into account both 3D point clouds and their 2D views, we create more robust pseudo-labels for self-training while simultaneously aligning the two data modalities. Furthermore, we use instance-level feature alignment to complement this objective.

\item We carefully engineer design elements from uni-modal self-training such as student-teacher networks for self-training, and masked-image-modeling for learning local features, to simultaneously accommodate multiple modalities. 

\item We demonstrate that the proposed joint pseudo-labels and feature alignment between images and point clouds enable each modality to benefit from one another's unique representations, leading to improved classification performance in each modality through label-free training. 

\end{itemize}

As shown in Figure \ref{fig:concept_figure_bargraph}, \textit{Cross-MoST} achieves respectively $+10.36\%$ and $+22.73\%$ improvement over zero-shot on the most widely used datasets; Scanobjectnn \cite{scanobjectnn} and Modelnet40 \cite{modelnet} respectively. It also achieves respectively $+11.53\%$ and $+4.25\%$ improvement over Self-training on the single point cloud modality, highlighting the impact of cross-modal learning.

It is important to note that Cross-MoST is orthogonal to works such as ULIP\cite{ULIP} which addresses self-supervised pre-training in images or point clouds, as well as works such as MUST\cite{MUST} which explores self-training on a single (image) modality. Advances in either or both domains will lead to even better Cross-modal Self-training paradigms.   
We present \textit{Cross-MoST} as a simple and effective solution for 3D classification that unlocks the potential of CLIP-like foundational models in practical scenarios where 3D scans and their corresponding 2D views are abundant, but the labels are scarce. 
We further demonstrate the effectiveness of our \textit{Cross-MoST} on 8 different versions of 4 popular 3D datasets, collectively representing synthetic and real-world 3D objects and 2D images, as well as real RGB, synthetic rendered, and depth-based 2D images. 

\section{Related work}

\textbf{Supervised Training:}
One approach to point cloud modeling involves projecting 3D point clouds into voxel or grid-based representation \cite{voxnet, pvrcnn}, followed by 2D or 3D convolutions for feature extraction. 
On the other hand, 
PointNet \cite{pointnet} and PointNet++ \cite{pointnet++} pioneered directly ingesting 3D point clouds and extracting permutation-invariant feature representations through shared MLPs. These networks have been widely used in various point cloud applications \cite{dgcnn, scfnet, paconv}. 
Similarly, PointNeXt \cite{pointnext} emerged as a lightweight version of PointNet++.  Point Transformer \cite{pointtrans} and PCT \cite{PCT} show improvements in the supervised training paradigm using transformer-based networks. 

\noindent\textbf{Self-Supervised Training:} self-supervised 3D pre-training methods \cite{sauder2019selfsupervised, poursaeed2020selfsupervised} utilize encoder-decoder architectures to first transform point clouds into latent representations, and then recover them into the original data form. Other works \cite{rao2020globallocal, PointContrast, fu2022posbert}  conducts self-supervised pre-training via contrastive learning. 
Inspired by the BERT \cite{bert} in the language domain and masked image modeling \cite{he2021masked, SimMIM}, several works have been proposed for masked point modeling for self-supervised learning \cite{pointbert, liu2022masked, pointmae, pointm2ae}.

\noindent\textbf{Multi-modal Pre-Training:}
Recent advancements in multi-modal contrastive learning have enabled CLIP \cite{CLIP} to perform robust and efficient multi-modal training with millions of image text pairs. CLIP has been extended to high-efficiency model training such as ALBEF \cite{albf}, cycle consistency \cite{cyclip}, and self-supervised learning \cite{slip}. 
PointClip \cite{pointclip, pointclipv2} and CLIP2Point \cite{clip2point} attempt to leverage the superior pre-trained knowledge of CLIP to improve 3D shape understanding by converting point clouds into multiple 2D views/depth maps.
ULIP \cite{ULIP} and CG3D \cite{cg3d} enable direct processing of point clouds using a separate 3D encoder, and leverage point cloud, image, and language triplets to align these three modalities together. 
Recent methods such as Uni3D\cite{zhou2023uni3d}, Vit-Lens\cite{lei2023vitlens}, and OpenShape\cite{liu2023openshape} scale up multi-modal pre-training using larger datasets.

\begin{figure*}[!t]
\centering
 \begin{minipage}{0.7\textwidth}
{\includegraphics[width=1.0\textwidth, trim={1.5cm 14cm 1.5cm 2.5cm},clip]{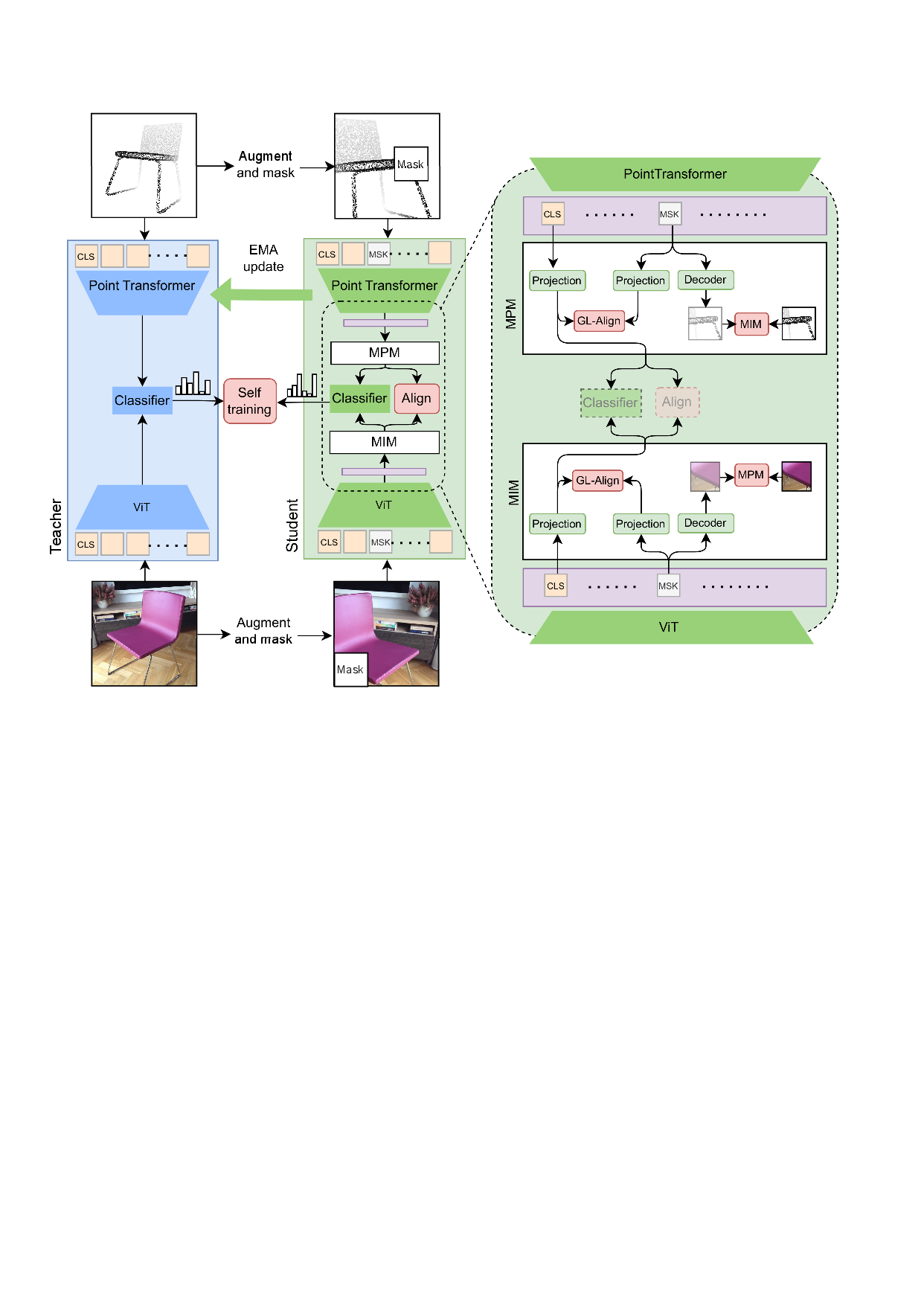}}
\end{minipage}
\hfill
  \begin{minipage}{0.28\textwidth}
\caption{Cross-modal Self-training for 3D point clouds and their corresponding 2D views. 
The teacher (\textcolor{blue}{blue}) weights are updated as an exponentially moving average of the student (\textcolor{green}{green}). The teacher generates joint pseudo-labels to allow cross-modal self-training. Our \textbf{MPM} and \textbf{MIM} modules inside the student model implement masked point and image modeling. \textbf{Align} represents the cross-modal feature alignment, whereas \textbf{GL-Align} within MIM and MPM modules represent global-local feature alignment to support masked modeling within each individual modality (image and pointcloud).}  
\label{fig:model_diagram}
 \end{minipage}
\end{figure*}

\noindent\textbf{Self-training for Semi-supervised Training:} 
Self-training improves the quality of features by propagating a small initial set of annotations to a large set of unlabeled instances, and has shown promising progress in domains including vision \cite{rethinking-self-training, SimMIM, better-self-training}, NLP \cite{self-training-neural}, and speech \cite{self-training-speech}. 
\cite{mean_teachers, FixMatch, remixmatch, MUST}.
Similar to \cite{MUST}, this work uses pseudo labels\cite{Lee2013PseudoLabelT} and applies consistency regulation \cite{FixMatch, temporal_ensembling} objective to encourage the model to output same predictions when perturbations are added to image/point cloud inputs, and guide the model to give sharp predictions with low entropy \cite{self-sup-entropy-minimization}.

We solve the problem of confirmation bias in self-training by combining pseudo-labels from complementary modalities to eliminate label noise and improve robustness
Furthermore, similar to \cite{mean_teachers, dino}, 
we model the teacher as an exponentially moving average of the student, thus improving the tolerance to inaccurate pseudo-labels.

\section{Cross-Modal Self-Training}
\label{method}

As shown in Figure.\ref{fig:model_diagram}, we operate in a unified embedding space for both image and point cloud branches.
We initialize the image encoder ViT with a CLIP pre-trained image encoder such that a common classifier could be initialized using CLIP text embeddings corresponding to the categories available in the training dataset. 
The point and image encoders, as well as the classifiers of student and teacher networks, are initialized identically.
An input training pair consists of a 3D point cloud, and an image representing the corresponding 3D object from a random view. 
The teacher network classifier generates the joint pseudo-label for the input image-point cloud pair by combining the two corresponding sets of classification logits. 
The same input pair is sent to the student network with heavy augmentations and masking to generate two sets of predictions using the student network classifier. 
Cross entropy loss is calculated between the joint pseudo labels and the student network predictions, and the weights of the classifier as well as both image and point cloud encoders are updated accordingly. Finally, the teacher network weights are updated as an exponentially moving average of the student.
Additionally, image and point mask modeling, as well as alignment losses complement our cross-modal self-training process as regularizers and additional supervision signals.

\subsection{Preliminaries}
\textbf{CLIP} \cite{CLIP} pre-trains an image encoder and a text encoder with a contrastive loss such that paired images and texts have high similarities in a shared embedding space. 
We denote the CLIP's image and text encoders as $h^{I}$ and $h^{S}$. The input image $X$ is divided into $K$ patches followed by a projection to produce patch tokens. Then a learnable [CLS] token is preppended to the input patch embeddings to create $\mathbf{X} \ =\{x_{cls} ,x_{1} ,x_{2} ,...x_{K}\}$ which is used as input to $h^{I}$, $h^{I}(\mathbf{X}) =\tilde{\mathbf{X}} =\left\{\tilde{x}_{cls} ,\tilde{x}_{1} ,\tilde{x}_{2} ,...\tilde{x}_{K}\right\}$. The output embedding of the [CLS] token is then normalized and projected to obtain the feature embedding of the image, $f^{I}\left(\tilde{x}_{cls}\right) =\tilde{x}$.

For zero-shot classification, each category $c$'s name is wrapped in several templates such as 
{\texttt{"a photo of a $\{$category$\}$"}, 
\texttt{"a picture a $\{$category$\}$"}} to produce $s_{c}$. These text descriptions are passed to the text encoder to yield the category-level normalized text embedding, $\tilde{s}_{c} =\ avg\left( h^{S}( s_{c})\right)$. During inference, the dot product between the text embeddings $\tilde{\mathbf{S}} =\left\{\tilde{s}_{i}\right\}_{c=1}^{C}$ and the image embedding yield prediction logits; $\tilde{x} .\tilde{\mathbf{S}} =p_{img}$

\noindent \textbf{Uni-modal self-training for images:} MUST \cite{MUST} proposes an EMA teacher-student setting to implement self-training on images using CLIP's zero shot prediction ability to generate pseudo-labels. It converts CLIP's non-parametric text embedding $\tilde{\mathbf{S}}$ into weights of a linear classifier $Q$ that takes as input the feature embedding of images; $Q\left(\tilde{x}\right) =p$. 
Both the EMA teacher and the student models are initialized with the same pre-trained weights, $\theta =\left\{\theta ^{I},\theta ^{Q}\right\}$, where $\theta ^{I}$ is the weights of CLIP visual encoder. The model teacher weights are updated at each iteration as: $\Delta =\mu \Delta +( 1-\mu ) \theta$. 
A batch $B$ of weakly augmented images is passed to the teacher model to yield a set of soft prediction logits $q_b$, which is converted to hard pseudo-labels as $\hat{q}_{b} =argmax_c( q_{b})$. The same batch of images is sent to the student model with strong augmentations to yield prediction logits $p_b$. Self-training loss is calculated as the cross-entropy $H$ between the predictions and the pseudo-labels that exceed a confidence threshold, $T$.
\begin{equation}
\mathcal{L}_{cls}=\frac{1}{B}\sum_{b=1}^{B}\mathbb{1}(\max( q_{b}) \geq T ) H(\hat{q}_{b} ,p_{b})
\end{equation}
This 
encourages the model to return the same predictions for perturbed inputs, while the student learns stronger representations as augmentations are applied to its input.

\subsection{Cross-Modal Self-training for images and pointclouds}
In this subsection, we elaborate the architectural elements and loss functions in Cross-MoST as shown in Figure.\ref{fig:model_diagram}.
\noindent \textbf{Pointcloud encoder:}
We denote an input point cloud as $Y$. Following the work of PointBert \cite{pointbert}, we first cluster the point cloud object into $K$ local patches or sub-clouds by applying the k-nearest-neighbor algorithm on a set of selected sub-cloud centers. Similar to patchifying in images, these sub-clouds contain only local geometric information, regardless of their original location. Next, they are passed through a pre-trained Point tokenizer from \cite{pointbert} to convert each sub-cloud into a point embedding, and a learnable [CLS] token is appended as follows; $\mathbf{Y} \ =\{y_{cls} ,y_{1} ,y_{2} ,...y_{K}\}$. This enables each point cloud object to be represented as a sequence of tokens and sent as input to a standard transformer encoder, $h^{P}$ as $ \tilde{\mathbf{Y}} =h^{P}(\mathbf{Y})=\left\{\tilde{y}_{cls} ,\tilde{y}_{1} ,\tilde{y}_{2} ,...\tilde{y}_{K}\right\}$. The output embedding of the [CLS] token $\tilde{y}_{cls}$ is normalized and projected to yield the point cloud feature embedding, $f^{P}\left(\tilde{y}_{cls}\right) =\tilde{y}$. 

Following the work of ULIP\cite{ULIP} we pre-train the point cloud encoder to learn a 3D representation space aligned with the image-text embedding space of CLIP. The pre-training is done on synthesized image-text-point cloud triplets from the CAD models of Shapenet\cite{shapenet} dataset. 

\noindent \textbf{EMA teacher-student setting:}
We use the non-parametric text embeddings $\tilde{\mathbf{S}}$ to initialize the classifier $Q$. Since both image and point cloud features are now projected to the same embedding space, \textit{the same classifier $Q$ is used to obtain prediction logits for both modalities.} $Q\left(\tilde{x}\right) =p_{img},~ Q\left(\tilde{y}\right) =p_{pcl}$. 
We use an ensemble of text prompts such as {\texttt{"a photo of a $\{$category$\}$"}, \texttt{"a 3D model of a $\{$category$\}$"}} and average them to initialize the classifier, to ensure sufficient zero-shot accuracy for both image and point cloud inputs.

Now, we characterize our model parameters as $\theta =\left\{\theta ^{I} ,\theta ^{P} ,\theta ^{Q}\right\}$, while the teacher is the EMA of the student model similar to MUST i.e.,  $\Delta =\mu \Delta +( 1-\mu ) \theta $. 
Our proposed self-training leverages cross-modal losses derived by joint pseudo-labels and feature alignment as well as regularization based on masked modeling and fairness. 

\noindent \textbf{Cross-Modal cross-entropy via joint pseudo-labels:} We create a training sample by pairing a point cloud object with an image showing the object from a random viewpoint. A set of weak augmentations are applied to the batch $B$ of image and point cloud pairs which is then passed through the teacher model to obtain two sets of soft prediction logits from each modality; $q_{b,img}, q_{b,pcl}$ which are used to derive modality-specific pseudo-labels $\hat{q}_{b,img} =\mathrm{argmax}_{c}( q_{b,img})$ and $\hat{q}_{b,pcl} =\mathrm{argmax}_{c}( q_{b,pcl})$.
Then, the teacher prediction for each sample $b$ that corresponds to the highest confidence between the two modalities in $\hat{q}_{b,img}$ and $\hat{q}_{b,pcl}$  is selected to assemble a set of joint pseudo-labels, $\hat{r}_{b}$.
The confidence score from the selected modality for each sample is defined as $r_{b}$; confidence score for the combined pseudo-label.
Following the work of FixMatch \cite{FixMatch} and MUST \cite{MUST}, we select the samples whose confidence exceeds a threshold $T$ to act as pseudo-labels.
(further details in Appendix \ref{appendix:joint_pseudolabels}).

The student model receives strongly augmented versions of the same image-point cloud pairs, yielding predictions $p_{b,img}, p_{b,pcl}$. 
We apply cross-entropy loss $H$ between the student predictions and the selected pseudo-labels.

\begin{equation}
\mathcal{L}_{cls} =\frac{1}{B}\sum _{b=1}^{B}\mathbb{1}( r_{b}  >T ) \{ H(\hat{r}_{b} ,p_{b,\ img}) + H(\hat{r}_{b} ,p_{b,\ pcl}) \}
\end{equation}

The pseudo-label selection through confidence thresholding indirectly
encourages the model to yield sharp predictions. Moreover, the combination facilitates cross-modal exchange of class-level feature knowledge while encouraging pseudo-label agreement between modalities.  

\noindent \textbf{Cross-modal feature alignment:} 
Extending the pertaining objective of CLIP and ULIP, we continue to enforce Feature alignment between image and point cloud pairs in the multimodal embedding space. Thereby we encourage images and point clouds representing one instance of reality to be embedded close to each other. This unsupervised objective complements the pseudo-supervised class-level discrimination implemented by self-training.

For a batch $B$ of training image-point cloud pairs, we calculate the cosine similarity between the feature embeddings of each image and point cloud
$\tilde{x}_{b}, \tilde{y}_{b}$, and 
maximize the similarity of image-point cloud embeddings of $B$ positive pairs, while minimizing the similarity of $B^2-B$ negative pairs.  We implement this objective by optimizing a symmetric cross-entropy over the similarity scores. 
\begin{equation}
\mathcal{L}_{align} =\sum _{( i,j)} -\frac{1}{2}\log\frac{\exp\left(\frac{\tilde{x}_{i}\tilde{y}_{j}}{\tau }\right)}{\sum _{b}\exp\left(\frac{\tilde{x}_{i}\tilde{y}_{b}}{\tau }\right)} -\frac{1}{2}\log\frac{\exp\left(\frac{\tilde{x}_{i}\tilde{y}_{j}}{\tau }\right)}{\sum _{b}\exp\left(\frac{\tilde{x}_{b}\tilde{y}_{j}}{\tau }\right)}
\end{equation}

\noindent \textbf{Fairness Regularization:} As shown by prior research \cite{wang2022debiased}, CLIP-like models are often biased towards certain classes, causing the pseudo-labels to further magnify such biases in the self-training settings. 
Therefore, we apply fairness regularization on both image and point cloud predictions separately during self-training by enforcing the following loss:
\begin{equation}
\mathcal{L}_{fair} =\ -\frac{1}{C}\sum _{c=1}^{C} \left(log(\overline{p}_{c,img}) + log(\overline{p}_{c,pcl} \right)
\end{equation}
where $\overline{p}$ denotes the batch-average prediction and $C$ is the number of class categories.

\noindent \textbf{Masked-Modeling MM:}
Masked image and point cloud modeling aim to learn specific local features at image-patch and point-subcloud levels respectively. It not only acts as a regularization to further reduce the noise in the pseudo labels but also as a complementary supervision signal from unlabeled data.

\noindent \textit{Masked-image reconstruction - \textbf{MIM:}}
We formulate the masked image modeling (MIM) objective as predicting the missing RGB values of masked-out patches using contextual information learned through attention. We train a simple linear decoder $g^{I}$ that takes output embedding $\tilde{x}_{m}$ of $m$th [MSK] token as input and reconstructs the masked-out RGB pixel values; $z_m =g^{I}\left(\tilde{x}_{m}\right)$ which are then compared with the ground truth patch to formulate the MIM loss. 
$$\mathcal{L}_{mim} =\frac{1}{MN}\sum _{m=1}^{M} \| z_m -\sigma _m \| _{1}$$  
$z_m, \sigma_m \in \mathds{R}^{N}$, $N$ is the number of RGB pixels in a patch, $\sigma_m$ is the  RGB values of the originally masked out patch.

\noindent \textit{Masked-point reconstruction - \textbf{MPM:}}
Representing each sub-cloud of a 3D spatial locality by a discrete token enables us to implement a masked point reconstruction objective. A selected set of tokens from $\mathbf{Y}$ is replaced by a learnable [MSK] token and passed as input to the point cloud encoder $h^P$. Random masking of patches from multiple locations makes the learning objective too easy given the richness of contextual information from neighboring patches in 3D point clouds. Therefore, we mask out a selected token, and m other tokens corresponding to sub-clouds in its spatial neighborhood resulting in a block-wise masking strategy \cite{pointbert}.  
We pass the output token through a linear decoder $g^{P}$; $w_m =g^{P}\left(\tilde{y}_{m}\right)$, and apply an $L_1$ loss between $w_m$ and corresponding the masked-out input token $y_m$. 

$$\mathcal{L}_{mpm} =\frac{1}{MN}\sum _{m=1}^{M} \| w_m -y_m \| _{1}$$ 
where $w_m ,\ y_m \in \mathds{R}^{N}$, and $N$ is the dimensionality of the point token. 
The model learns to predict missing geometric structures of a point cloud given its neighboring geometries, encouraging local feature representation learning.

\noindent \textbf{Global-local alignment:}
Both MIM and MPM encourage each branch to learn rich local semantics. 
Then we transfer this local feature information to the object-level embedding space by aligning local and global features within each modality. Specifically, we project the output embeddings of all [MSK] tokens to the image/point cloud feature embedding space using $f^I$ and $f^P$, and calculate a global-local feature alignment loss \textit{within} each modality.

$$\mathcal{L}_{lg-align} =\frac{1}{M_{img}}\sum _{m=1}^{M_{img}} \| \tilde{x} -u_{m} \| _{2}^{2} +\frac{1}{M_{pcl}}\sum _{m=1}^{M_{pcl}} \| \tilde{y} -v_{m} \| _{2}^{2}$$ where $u_{m} =f^{I}\left(\tilde{x}_{m}\right)$, $\tilde{x}_{m}$ is the output embedding of the $m^{th}$ masked image token, and where $v^{m} =f^{P}\left(\tilde{y}_{m}\right)$, $\tilde{y}_{m}$ is the output embedding of the $m^{th}$ masked point token.

\section{Experimental Protocols}
\label{Experiments}

\textbf{Datasets details:} \textbf{Shapenet} \cite{shapenet} consists of textured CAD models of 55 object categories. 
We use Shapenet to pre-train the point cloud branch for better self-training initialization.
\textbf{ModelNet40} \cite{modelnet} is a synthetic dataset of 3D CAD models containing 40 categories. 
We pair 2D renderings of CAD models with the point clouds to create \textbf{Modelnet40} and \textbf{ModelNet10} (a subset of 10 common classes). We follow the realistic 2D views generated using \cite{pointclipv2} to generate the dataset \textbf{ModelNet40-d} (depth). 
\textbf{Redwood} \cite{choi2016large} is a dataset of real-life high-quality 3D scans and their mesh reconstructions. 
We randomly sample 20 frames from the RGB videos of each object scan and use them in our image encoder. 
\textbf{Co3D} \cite{reizenstein2021common} is a large-scale dataset of real multi-view images capturing common 3D objects, and their SLAM reconstructions. We sample 20 GRB images per object for our image encoder.
For \textit{Scanobjectnn} \cite{scanobjectnn} with real 3D point cloud scans, we report results on 3 different versions; \textbf{Sc-obj} - scans of clean point cloud objects, \textbf{Sc-obj withbg} - scans of objects with backgrounds, and \textbf{Sc-obj hardest} - scans with backgrounds and additional random scaling and rotation augmentations. Multiview images for all versions of scanobjectnn are generated using realistic 2D view rendering from point clouds\cite{pointclipv2}.
Detailed descriptions of the sizes, number of categories, and pre-processing applied to each dataset are provided in Appendix \ref{appendix:datasets}.

\noindent \textbf{Implementation details:} We use ViTB/16 as the image encoder, and a standard transformer \cite{vaswani2017attention}  with multi-headed self-attention layers and FFN blocks as our point cloud encoder. 
We use AdamW \cite{adam} optimizer with a weight decay of 0.05. 
The batch size is 512, and the learning rate is scaled linearly with batch size as (lr= base\_lr*batchsize/256). We used 4 V100 GPUs for training. Further details are in Appendix \ref{appendix:implementation_details}.

\noindent \textbf{Image encoder:} 
We use a ViT-B/16 model pretrained by \cite{CLIP} for the image branch. 
After resizing to the side $224\times224$, random cropping is applied as a weak augmentation on the input to the teacher model to generate pseudo-labels. 
Stronger augmentations; RandomResizedCrop+Flip+RandAug \cite{cubuk2019randaugment} are applied to the inputs to the student model.
We implement a patch-aligned random masking strategy where multiple image patches are randomly masked with a fixed ratio of 30\%. 

\begin{table*}[!t]
\small \centering
\setlength{\tabcolsep}{1.6mm}{
\scalebox{0.9}[0.9]{
\begin{tabular}
{llcccccccc}
\toprule
Method & \textbf{Datasets $\rightarrow$} & {ModelNet10} & {ModelNet40} & {ModelNet40-d}  & {Redwood} & {Co3d} & {Sc-obj} & {Sc-obj withbg} & {Sc-obj hardest} \\
\midrule
{Baseline} & Image & 55.00 & 54.00 & 23.18 & 85.71 & 90.30 & 19.11 & 16.70 & 13.12 \\
{Zeroshot} & Image* & 65.50 & 56.25 & 30.31  & 85.71 & 94.01 & 26.16 & 19.28 & 13.85 \\
 & Point cloud & 75.50 & 58.75 & 57.74  & 55.95 & 13.20 & 46.99 & 42.86 & 31.57 \\
 \midrule
{Baseline} & Image & 78.00 & 73.13 & 35.58  & 86.91 & 91.08 & 23.24 & 20.65 & 18.25 \\
{Self-training} & Image* & 85.50 & 78.88 & 46.80  & 91.67 & 92.44 & 27.54 & 28.23 & 23.87 \\
 & Point cloud & 85.50 & 69.13 & 61.26  & 63.10 & 16.90 & 47.33 & 49.57 & 37.47 \\
 \midrule
{Cross-modal} & Image & 86.50 & 79.50 & 52.92  & 88.10 & 93.15 & 52.84 & 49.57 & 40.60 \\

{self-training} & Image* & 89.50 & 82.75 & 62.89 &  \textbf{94.05} & \textbf{94.22} & \textbf{58.86} & \textbf{55.94}& \textbf{44.14}\\
& Point cloud & \textbf{90.00}  & \textbf{83.13}  & \textbf{66.17}  & 75.00 & 83.45 & 48.02 & 51.46 & 41.64 \\
    \bottomrule
\end{tabular}%
}}
\caption{\small We evaluate the classifier on 2D views/images and 3D point clouds separately. Image* indicates the performance of the classifier averaged over all rendered views. 
For each dataset, we highlight the highest achieved accuracy among the three evaluation settings.
Cross-modal self-training consistently improves over Zeroshot and self-training baselines for both images and point clouds highlighting the potential of Cross-modal learning.}
\label{tab:main_results}
\end{table*}
\begin{table*}[]
\small \centering
\setlength{\tabcolsep}{1.6mm}{
\scalebox{0.9}[0.9]{
\begin{tabular}
{llcccccccc}
\toprule
 Method & \textbf{Datasets $\rightarrow$} & ModelNet10 & ModelNet40 & ModelNet40-d  & Redwood & co3d & Sc-obj & Sc-obj withbg & Sc-obj hardest \\
\midrule
Cross-modal & Image & 86.50 & 79.50 & 52.92 & 88.10 & 93.15 & 52.84 & 49.57 & 40.60 \\
Self-training & Image* & 89.50 & 82.75 & 62.89 & \textbf{94.05} & \textbf{94.22} & \textbf{58.86} & 55.94 & \textbf{44.14} \\
 & Point cloud & 90.00 & \textbf{83.13} & 66.17 & 75.00 & 83.45 & 48.02 & 51.46 & 41.64 \\ \midrule
Without L-Align & Image & 86.50 & 78.38 & 53.57 & 88.10 & 88.30 & 48.71 & 48.02 & 39.24 \\
 & Image* & \textbf{91.00} & 83.00 & 62.24 & \textbf{94.05} & 90.30 & 55.25 & 55.25 & 43.13 \\
 & Point cloud & \textbf{91.00} & 81.75 & 65.40 & 75.00 & 81.39 & 48.19 & 50.60 & 39.59 \\ \midrule
Without MM & Image & 87.50 & 80.13 & 57.66 & 83.33 & 81.74 & 50.26 & 52.32 & 40.94 \\
 & Image* & \textbf{91.00} & 82.88 & 66.37 & 88.10 & 83.81 & 54.39 & \textbf{59.55} & 43.72 \\
 & Point cloud & 89.50 & 81.75 & \textbf{67.10} & 73.81 & 79.53 & 49.91 & 55.94 & 42.05 \\ \bottomrule
\end{tabular}%
}}

\caption{Ablations on all datasets.
\textbf{Without L-Align} refers to ablating the cross-modal feature alignment by removing $L_{align}$.
\textbf{Without MM} refers to ablating the masked-modeling in both branches; image and point cloud by removing both $L_{mim}$ and $L_{mpm}$.}

\label{tab:all-dataset-ablation}
\end{table*}
\begin{table*}[t!]
    \centering
    \begin{minipage}{.38\textwidth}
    \setlength{\tabcolsep}{8pt}{
    \resizebox{1\linewidth}{!}{
    \begin{tabular}{cccc|ccc}
    \toprule
 {Align}  & {Comb} & {MM} & {Init.} & {Image} & {Image*} & {Pointcloud} \\
            \midrule        
\xmark & \cmark  & \cmark & \cmark & 78.38 & 83.00  & 81.75 \\

\xmark & \xmark  & \cmark & \cmark & 73.13 & 78.88  & 69.13 \\

\cmark & \xmark  & \cmark & \cmark & 73.75 & 77.38  & 72.63 \\

\cmark & \cmark  & \xmark & \cmark & 80.13 & 82.88  & 81.75 \\
\cmark & \cmark  & \cmark & \xmark & 73.00 & 78.63  & 21.88  \\
\cmark & \cmark  & \cmark & \cmark & 79.50 & 82.75 & \textbf{83.13} \\
\bottomrule
\end{tabular}
}}
\caption{\small \textbf{Aign}- Cross-modal feature alignment, \textbf{Comb}- Joint pseudo-label, \textbf{MM}-Image and Point masked modeling, and \textbf{Init.}- point cloud encoder initialization using ULIP pre-training. Results are reported on Modelnet40.
}
\label{tab:Ablation-main}
\end{minipage}
\hfill
\begin{minipage}{0.33\textwidth}
    \centering
\setlength{\tabcolsep}{8pt}{
\resizebox{1\linewidth}{!}{
\begin{tabular}
{l|cccc}
\toprule
{Pseudo-labels} & {Image} & {Image*} & {Point cloud} \\
\midrule
No Comb. & 73.75 & 77.38 \ & 72.63\\
Image only & 72.50 & 76.88 & 78.38  \\
Point cloud only & 71.63 & 74.38 & 70.50 \\
Random  & 78.25 & 81.88  & 81.88  \\
Our (Joint) & 79.50 & 82.75 & \textbf{83.13} \\
\bottomrule
\end{tabular}
}}
\caption{\small Effect of pseudo-labels on self-training for modelnet40. \textbf{No comb.} does self-training on individual modalities using its own pseudo-labels. \textbf{Only image/pointcloud} adapts the pseudo-label from one of the modalities for both.}
\label{tab:different-combination-methods}
\end{minipage}
\hfill
\begin{minipage}{0.25\textwidth}
    \centering 
\setlength{\tabcolsep}{8pt}{
\resizebox{1\linewidth}{!}{
\begin{tabular}
{c|ccc}
\toprule
{Views} & {Image} & {Image*} & {Pointcloud} \\
\midrule
1 & 76.75 & 76.75  & 78.50  \\
2 & 78.38 & 80.75  & 80.13  \\
4 & 78.50 & 82.38  & 80.50  \\
8 & 78.88 & 82.00  & 81.25  \\
12 & 79.50 & 82.75 & \textbf{83.13} \\
\bottomrule
\end{tabular}
}}
\caption{\small Increasing the number of 2D rendered views per point cloud improves the performance of our cross-modal self-training. Results are reported on Modelnet40.}
\label{tab:nviews_ablation}
\end{minipage}
\vspace{-15pt}
\end{table*}

\noindent \textbf{Point cloud encoder:} We used Shapenet \cite{shapenet} dataset and its rendered 2D views from \cite{ULIP}
to pre-train the point cloud encoder and the projection layers $f^P$, $f^S$, and $f^I$.
Pre-training is done for 250 epochs with a learning rate of $10^{-3}$ with AdamW optimizer with a batch size of 64. 
We divide each point cloud into 64 sub-clouds and
use a Mini-Pointnet to extract embeddings of each sub-cloud, followed by a 
pointbert\cite{pointbert} encoder to convert each sub-cloud into point embeddings. 
Rotation perturbation and random scaling
are applied on the input point cloud to the teacher model to generate pseudo-labels. 
Stonger augmentations; random cropping, input dropout, rotate, translate, and scaling 
are applied to the input to the student model.  Random masking is applied to 30\% to 40\% of the point embeddings. 

\subsection{Results}
\label{subsec:Results}
We perform experiments on 4 datasets with point clouds and associated images, which span different types of 3D point clouds; \textit{Modelnet}- sampled from CAD models, \textit{Redwood}- real 3D scans, and \textit{Co3D}- SLAM reconstructed point clouds. We also use different types of images; \textit{Modelnet}- 2D rendered CAD models, \textit{Redwood, Co3D}- real images, and \textit{Scanobjectnn}- realistic depth renderings from point clouds.
Table \ref{tab:main_results} shows the results of our proposed cross-modal self-training. 

\noindent \textbf{Baselines}: 1) \textit{Baseline Zeroshot}- we used ULIP \cite{ULIP} trained with CLIP \cite{CLIP} initialization. We use the embeddings of text prompts {\texttt{"a photo of a $\{$category$\}$"}, 
\texttt{"a 3D model of a $\{$category$\}$"}} to initialize the classifier and directly evaluate the classification.
2) \emph{Baseline self-training \cite{MUST}}- 
We removed all cross-modal modules and cross-modal losses, and applied self-training \emph{individually} on 2D and 3D branches using their own pseudo-labels without any cross-modal combination. For both baselines, ULIP pretraining was done on the point cloud branch to provide consistent initialization.

\noindent \textbf{Evaluation:} The same classifier operates on both images and point cloud embeddings. To emulate real-life scenarios where both image and point cloud data are simultaneously available for a test sample, as well as to show how both modalities benefit from each others' unique knowledge, we report the classification accuracies for \textit{Image} and \textit{Point cloud} encoders separately. \textit{Image*} is calculated by using the average of image embeddings of all 2D views corresponding to the test object, hence is often higher than \textit{Image} due to richer information from multiple views. Datasets such as Co3D have very noisy and occluded point clouds, but are accompanied by high-quality RGB images; leading to higher accuracy on \textit{Image} and \textit{Image*}. We report the accuracies of both branches to demonstrate the cross-modal learning impacts the two modalities. 

As shown in Table \ref{tab:main_results}, we substantially improve upon the baselines on all datasets. 
Especially, comparison between Baseline Self-training (individual self-training on each branch without any cross-modal label or feature exchange) and Cross-modal self-training for image and point cloud branches suggests that our proposed method enables both modalities to learn from each other

An important result is that for datasets such as \textit{Modelnet40-d} whose zero-shot accuracies on the point cloud branch are higher than that of the image (57.74\% and 30.31\% respectively), Cross-modal self-training significantly improves point cloud branch accuracy above baseline self-training (from 61.26\% to 66.17\%). This effect is even more pronounced in the variants of \textit{Scanobjectnn} resulting in an even higher performance on \textit{Image*} compared to \textit{Point cloud}.
Similarly, for datasets such as \textit{Redwood} whose zero-shot accuracies on image branch are higher than that of point clouds (85.71\% and 55.95\% respectively), Cross-modal self-training significantly improves image branch accuracy above baseline self-training (from 91.67\% to 94.05\%).
This shows that \textit{even with low zero-shot performance, unique knowledge of the 3D branch due to their rich geometric details and that of the image branch due to large-scale CLIP pertaining can provide strong complementary training signals to the other modality.}

\noindent \textbf{Comparisons with SOTA:}
\begin{table}[t!]
\centering 
\setlength{\tabcolsep}{1mm}{
\resizebox{0.9\linewidth}{!}{
\begin{tabular}
{l|c}
\toprule
{Method} & {Modelnet40 (top1) accuracy} \\
\midrule
Openshape\cite{liu2023openshape}-Pointbert  & 70.3  \\
VIT-LENS-Datacomp-L14\cite{lei2023vitlens}  & 70.6 \\
ULIP-Pointbert\cite{ULIP}  & 60.4  \\
ULIP-Pointbert with Cross-MoST & \textbf{83.13 }\textcolor{blue}{(+22.73)} \\
\bottomrule
\end{tabular}
}}
\caption{\small Reported performance of state-of-the art Zero-shot models trained on Shapenet\cite{shapenet} compared with proposed Cross-MoST}
\label{tab:sota}
\vspace{-15pt}
\end{table} 
In Table.\ref{tab:sota}, we report the performance of state-of-the-art pre-training methods on modelnet40.
To preserve consistency, we only include the methods that use Shapenet for pre-training. 
However, it is important to distinguish between open-vocabulary and self-training settings. Although no labels are used in the process, self-training can be framed as a further adaptation of an open-vocabulary model for a specific set of categories. 
Another implication is that 
recent improvements in pre-training \cite{lei2023vitlens,liu2023openshape} could further improve the performance of cross-modal self-training by providing even better initialization than ULIP\cite{ULIP}.

\subsection{Ablative Analysis}

\begin{table*}[!t]
    \centering
    \begin{minipage}{0.48\textwidth}
        \centering
        
    \includegraphics[trim={0cm 1cm 0 0},clip,width=0.8\linewidth]{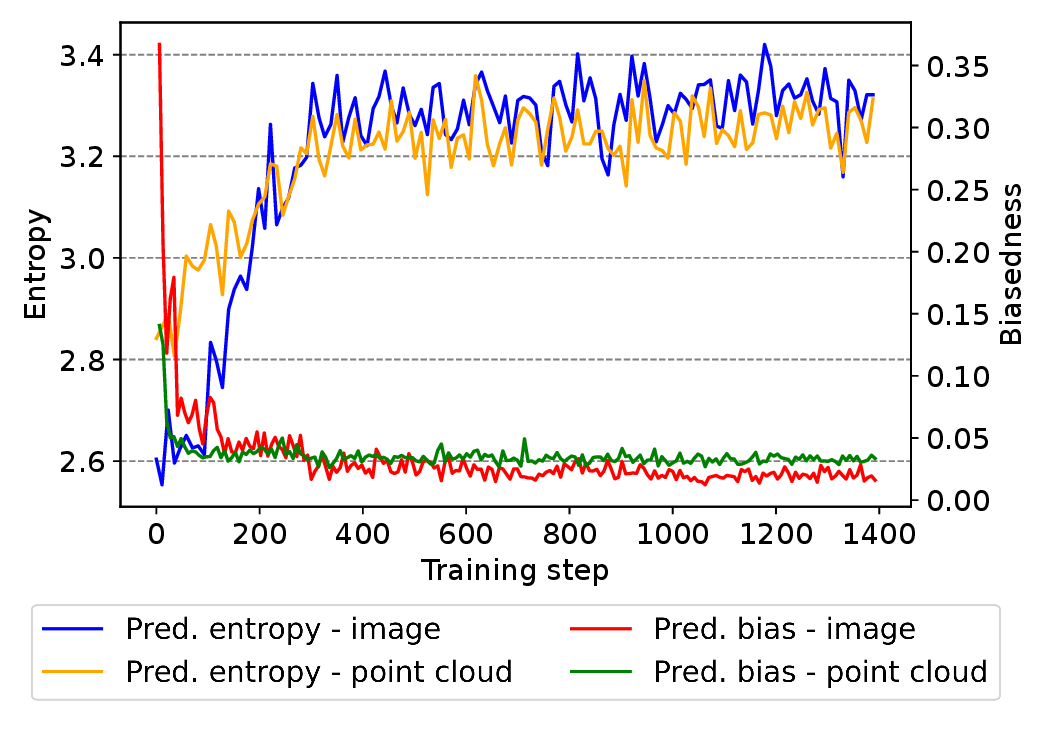}
  \captionof{figure}{\small As the training progresses, biasness towards certain classes is significantly reduced in both branches. Predictions on each branch become more sharp, as indicated by increasing entropy. (modelnet40)}
  \label{fig:biasedness_entropy} 
  
    \end{minipage}
    \hfill
    \begin{minipage}{0.48\textwidth}
          \centering

    \includegraphics[trim={0cm 1.5cm 0 0},clip,width=0.8\linewidth]{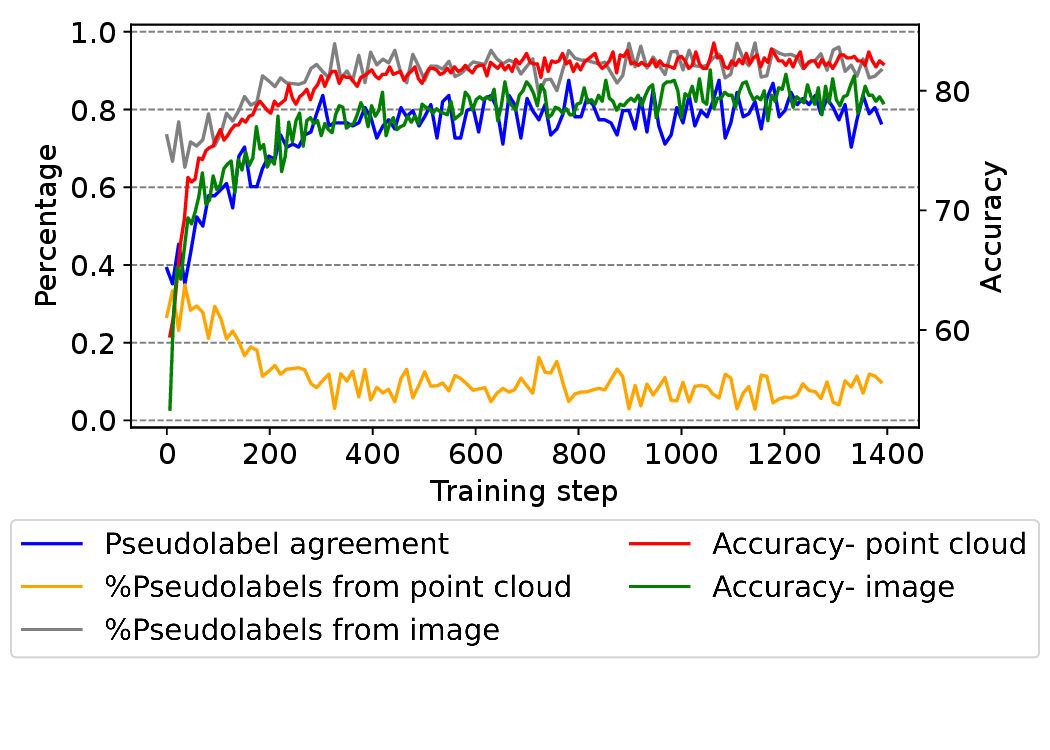}
  \captionof{figure}{\small The percentage of pseudo-labels selected from each modality for combined self-training. The agreement between pseudo-labels increases as our training progresses. (modelnet40)}
  \label{fig:pseudolabel_percent}

    \end{minipage}
\vspace{-15pt}
\end{table*}

\begin{table}[t!]
\centering
\setlength{\tabcolsep}{1mm}{
\resizebox{0.9\linewidth}{!}{
\begin{tabular}
{llccc}
\toprule
 & \textbf{Class Type $\rightarrow$} & {All classes} & {Medium classes} & {Hard classes} \\
\midrule
{Baseline} & Image & 54.00 & 47.50 & 44.71 \\
 {Zeroshot} & Image* & 56.25 & \textbf{49.55} & \textbf{47.65} \\
 & Point cloud & \textbf{59.50} & 42.73 & 36.76 \\
\midrule
{Self-training} & Image & 73.13 & 67.73 & 66.18 \\
{Baseline} & Image* & 78.88 & 73.64 & 72.06 \\
 & Point cloud & 69.25 & 57.27 & 52.06 \\
\midrule
{Cross-modal} & Image & 79.50 & 72.73 & 72.65 \\
{self-training}  & Image* & 82.75 & \textbf{77.27} & \textbf{77.65} \\
 & Point cloud & \textbf{83.13} & 76.82 & 76.47 \\
\bottomrule
\end{tabular}
}}
\caption{\small Accuracy on unseen classes for Modelnet40. Cross-modal self-training significantly outperforms self-training in \textit{medium} and \textit{hard} classes, especially in the point cloud branch.}
\label{tab:modelnet40 medium hard}
\vspace{-10pt}
\end{table}

\textbf{Effect of proposed objectives:} In Table \ref{tab:Ablation-main}, we ablate the main components of the proposed cross-modal self-training setting for Modelnet40, and illustrate their contribution to the final architecture.
The best performance is achieved when all the components are used in combination. 
It is also important to note that pretraining the point cloud encoder on even a limited dataset such as Shapenet dramatically improves the performance on both point cloud and image branches after Cross-modal self-training.

In Table \ref{tab:all-dataset-ablation}, we report the results of ablating Cross-modal feature alignment and Masked-modeling and confirm that these design elements lead to a clear improvement of performance in a majority of datasets. 
As we qualitatively compare in Appendix \ref{appendix:datasets}, the quality, hence the difficulty of these datasets varies dramatically in both modalities. 
Masked modeling is more advantageous to datasets (such as Redwood and Scanobjectnn) with point clouds with large distribution shifts from Shapenet due to obfuscations and heavy augmentations.
Furthermore, in Co3D image branch, ablating masked modeling leads to a degradation in performance indicating the importance of its regularization effects.

\noindent \textbf{Effect of pseudo-label combination:} Table \ref{tab:different-combination-methods} compares the performance of the model with different approaches to derive pseudo-labels.
\textit{Random} refers to randomly picking a prediction from either branch image or point cloud to act as the pseudo-label for a given input pair. 
Cross-modal learning significantly improves the performance on both image and point cloud branches even with a random combination of pseudo-labels. 
The proposed score-based method
further improves accuracy by using the most confident predictions between the modalities to act as the pseudo-label.

\noindent \textbf{Effect of the number of 2D views:} In our experiments with ModelNet40 \cite{su2015multiview} we have 12 2D views for each pointcloud object. The ablation results in Table \ref{tab:nviews_ablation} indicate that object understanding benefits from using multiple different views to extract more detailed visual understanding. This improvement is also reflected in the point cloud branch due to the cross-modal training.

\noindent \textbf{Training Analysis:} By thresholding the score of teacher predictions from each branch and combining them to generate joint pseudo-labels, we implicitly encourage the model to give sharp predictions. We calculate \textit{prediction entropy} as the \textit{KL divergence between a uniform distribution and the softmax predictions} of each branch to quantify this behavior. Figure \ref{fig:biasedness_entropy} illustrates how the entropy of predictions increases with self-training.
CLIP models are known to result in predictions biased towards certain classes \cite{wang2022debiased}. This hinders the ability to self-train since such biases can be further amplified \cite{confirmation-bias}. Figure \ref{fig:biasedness_entropy} shows how our regularization and cross-modal learning objectives discourage this confirmation bias as training progresses. The \textit{biasedness} is calculated as \textit{KL divergence between a uniform distribution and the class distribution of the predictions for a balanced test set} (further details on the calculation of entropy and biasedness are in Appendix \ref{appendix:training_trends_equations}).

Figure \ref{fig:pseudolabel_percent} shows the percentage of pseudo-labels picked from each modality for the combined self-training. Although the accuracy of each individual branch is comparable, the modal steers itself to pick more pseudo-labels from the image branch as training progresses.

\vspace{-10pt}

\paragraph{Self-training on unseen classes:} Certain classes of ModelNet40, have been already introduced to the model during supervised pre-training of the point cloud encoder by Shapenet55.
For a fairer comparison of zeroshot and label-free classification performance, therefore we evaluate our model on 2 other splits of ModelNet40 as proposed by \cite{ULIP}- \textit{medium} and \textit{hard}, with non-overlapping object classes (further details of these splits in Appendix \ref{appendix:datasets}). Results in table \ref{tab:modelnet40 medium hard} show that cross-modal self-training significantly improves the accuracy of hard and medium classes.

\section{Conclusion} 
In this paper, we proposed a simple framework to adapt an open-vocabulary 3D vision model to downstream classification without using any labels.
The core of the proposed approach is to enable cross-modal self-training by leveraging pseudo-labels from point clouds and their corresponding 2D views and additionally aligning their feature representations at the instance level. 
The proposed method is orthogonal to pretrained foundational models and the quality of 2D images that provide complementary information. Therefore, improvements in pre-trained foundational models or the quality of 2D views/renderings of point clouds will further improve the results of self-training in our proposed framework.
Our work highlights how the rich knowledge of CLIP-based models can be adapted to better understand 3D realities even in the absence of class-level labels.


\twocolumn[
\begin{@twocolumnfalse}
\begin{center}
\textbf{{\Large Appendix- Supplementary Materials}\\[0.5em]
{\large Cross-modal Self-training: Aligning Images and Pointclouds to learn Classification without Labels}
}
\end{center}\vspace{1em}
\end{@twocolumnfalse}]
The following section contains supplemental information and encompasses more implementation details, a comparison of datasets, and further analysis of discriminative features learned by the proposed cross-modal self-training. 
The contents are organized in the order listed below.
\begin{itemize}
    \item Cross-modal joint pseudo-labels (Appendix~\ref{appendix:joint_pseudolabels})
    \item Additional implementation details (Appendix~\ref{appendix:implementation_details})
    \item Datasets
    (Appendix~\ref{appendix:datasets})
    \item Training trends
    (Appendix~\ref{appendix:training_trends_equations})
    \item Analysis of feature embeddings (Appendix~\ref{appendix:tsne_embeddings})
\end{itemize}

\section{Cross-modal joint pseudo-labels} \label{appendix:joint_pseudolabels}
We generate a training sample by pairing a point cloud object with an image showing the object from a random viewpoint. A set of weak augmentations are applied to the batch $B$ of image and point cloud pairs which is then passed through the teacher model to obtain two sets of soft prediction logits from each modality; $q_{b,img}, q_{b,pcl}$
Then, the teacher prediction for each sample that corresponds to the highest confidence between the two modalities is selected to generate a common set of pseudo-labels for both image and point cloud modalities as follows:
\begin{align*}
\hat{r}_{b} =\{\mathbb{1}\left( \max( q_{b,img}) \geq \max( q_{b,pcl})\right) \hat{q}_{b,img} \\ \cup\; \mathbb{1} \left( \max( q_{b,img}) < \max( q_{b,pcl}) \right) \hat{q}_{b,pcl} \ \}, 
\label{eq:score-based-pseudolabel}
\end{align*}
where $\hat{q}_{b,img} =\mathrm{argmax}_{c}( q_{b,img})$ and $\hat{q}_{b,pcl} =\mathrm{argmax}_{c}( q_{b,pcl})$.
The confidence scores for the combined pseudo-labels are denoted by
$$r_{b} =\max_{b}\{\max_{c}( q_{b,img}) ,\max_{c}( q_{b,pcl})\}$$ where $r_{b} \in \mathds{R}^{B}$. 

\section{Implementation details} \label{appendix:implementation_details}
We use ViTB16 as the image encoder, and a standard transformer \cite{vaswani2017attention}  with multi-headed self-attention layers and FFN blocks as the point cloud encoder. The [MSK] tokens, projection layers, and decoder layers are randomly initialized, and the classifier is initialized with the text embedding as explained in \ref{method}. Then they are fine-tuned together with the encoders in the EMA teacher-student setting.
We use AdamW \cite{adam} optimizer with a weight decay of 0.05. We apply a cosine learning rate scheduler and similar to \cite{bao2022beit, he2021masked}, we apply layer-wise learning rate decay of 0.65. The batch size is 512, and the learning rate is scaled linearly with the batch size as (lr= base\_lr*batchsize/256). We used 4 V100 GPUs for training. 

\textbf{Image encoder:} We initialize image encoder with \cite{CLIP} weights. We use a ViT-B/16 model pretrained by \cite{CLIP} for the image branch, containing 12 transformer blocks with 768 dimensions. The model receives input images of size $224\times224$. After resizing, random cropping is applied as a weak augmentation on the input to the teacher model to generate pseudo-labels. A set of stronger augmentations; RandomResizedCrop+Flip+RandAug \cite{cubuk2019randaugment} is applied to the input to the student model.
We implement a patch-aligned random masking strategy where multiple image patches are randomly masked with a fixed ratio of 30\%. 

\textbf{Point cloud encoder:} We used Shapenet \cite{shapenet} dataset and its rendered 2D views from \cite{ULIP}
to pre-train the point cloud encoder and the  projection layers $f^P$, $f^S$, and $f^I$ while the image and text encoders are kept frozen. Training is done for 250 epochs with a learning rate of $10^{-3}$ with AdamW optimizer with a batch size of 64. We divide each point cloud into 64-point patches each containing 32 points. Point centers for clustering local point patches are selected using farthest point sampling (FPS). 
Then the locations of corresponding point centers are subtracted from the local patches such that they contain only the local geometric information irrespective of its original position. 
We use a Mini-Pointnet to extract point embeddings of each sub-cloud, followed by a sharpened-pre trained  encoder from \cite{pointbert} to convert each sub-cloud into point embeddings. The center locations of each sub-cloud are passed through an MLP to encode a positional embedding and are appended to the point embeddings, before passing to the transformer. The depth of the transformer is set to 12, the feature dimension to 384, and the number of heads to 6. During point encoder pre-training as well as self-training, the  encoder is kept frozen.
After normalizing, weak augmentations such as rotation perturbation and random scaling in the range of 90\%~110\% are applied on the input point-cloud to the teacher model to generate pseudo-labels. Stonger augmentations; random cropping, input dropout, rotate, translate, and scaling in the range of 50\%~200\% are applied to the input to the student model.  Random masking is applied to 30\% to 40\% of the point embeddings.

\section{Datasets}\label{appendix:datasets}

\begin{figure*}[h!]
\centering
{\includegraphics[width=0.8\textwidth]{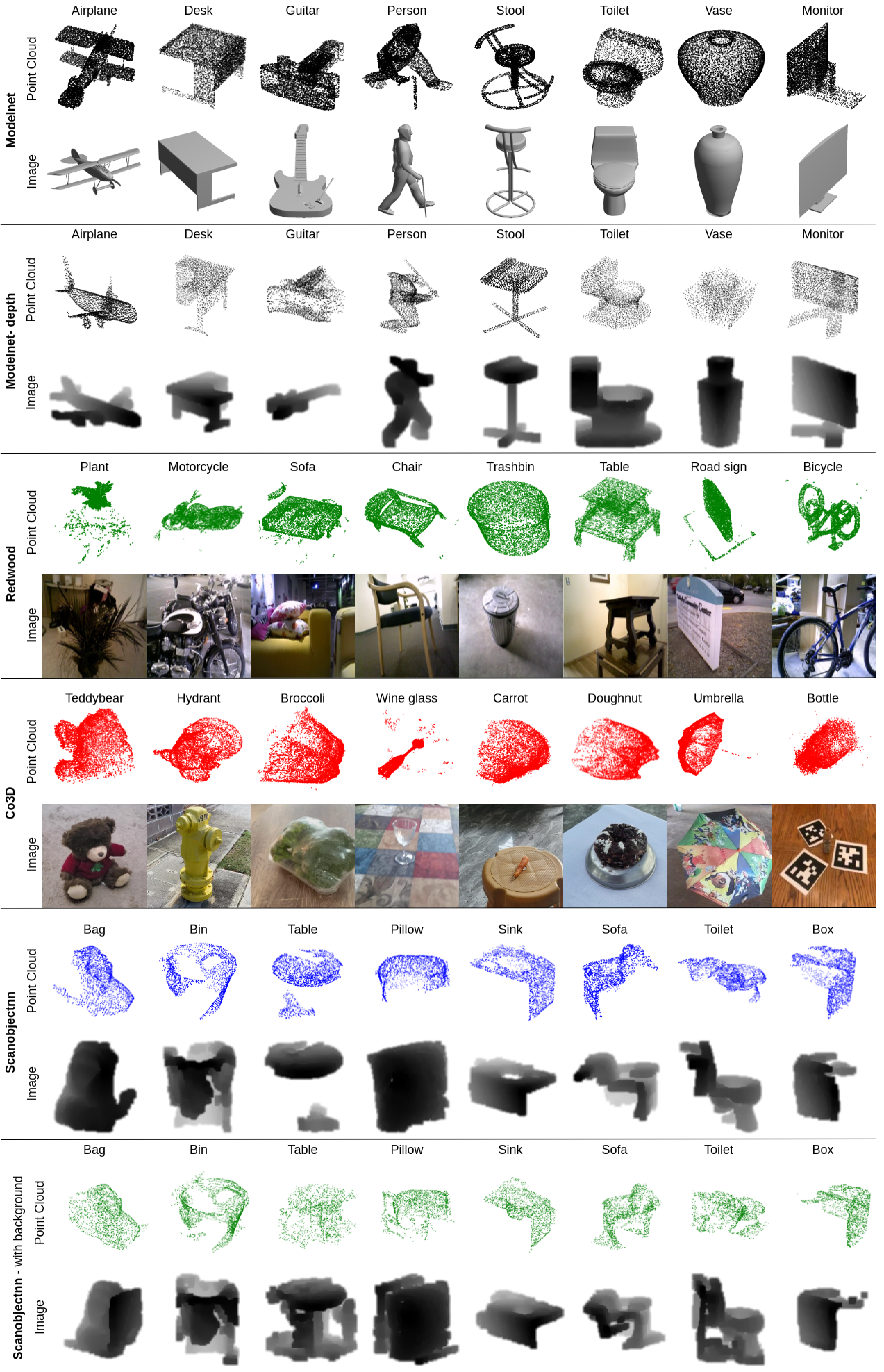}}
\caption{\small Qualitative comparison of datasets.}  
\label{fig:qualitative-dataset}
\end{figure*}

\paragraph{Shapenet \cite{shapenet}} consists of textured CAD models of 55 object categories. We uniformly sample points from each object mesh to create the pointclouds, after which they are normalized to fit a unit sphere. Following the work of \cite{ULIP}, we render RGB images for each object from different viewpoints. We use this dataset to pretrain the pointcloud branch to provide a better initialization for self-training.

\paragraph{ModelNet40 \cite{modelnet}} is a synthetic dataset of 3D CAD models containing 40 categories. The pointclouds are sampled from the object mesh and normalized. We use the work of \cite{su2015multiview} to get realistic 2D renderings of the CAD models. We pair these renderings with the pointclouds to create Modelnet40 and ModelNet10 (A subset of 10 common classes from \cite{modelnet}). We follow the realistic 2D views generated using \cite{pointclipv2} to generate the dataset ModelNet40-d (depth). 

\paragraph{Redwood \cite{choi2016large}} is a dataset of real-life high-quality 3D scans and their mesh reconstructions. 
However, these object meshes also contain points from the floor plane and surrounding objects. Unlike CAD models in \cite{shapenet, modelnet}, Redwood scans are not axis-aligned. For these models to be consistent with such datasets, we run a RANSAC plane detection to identify the floor plane and then rotate the orientation of the object to match the 3D coordinate axes. We then removed the points from surrounding objects/noises by manually cropping every scan.
We randomly sample 20 frames from the RGB videos of each object scan and use them in our image branch.

\paragraph{Co3D \cite{reizenstein2021common}} is a large scale dataset of multiview images capturing common 3D objects. These 2D views have been used to reconstruct 3D point clouds representing each object. Following the work of the original authors, we filter out only the accurate 3D point clouds for our experiments. Being SLAM-based reconstructions, these point cloud objects are also not axis-aligned. We implement a rough correction to the orientations by calculating the principal component directions based on the point densities of the point cloud and rotating the object to align it with the gravity axis in the 3D coordinates. 
However, this is a very challenging point cloud dataset due to differences in orientation, obscured parts of objects, and surrounding noises.
We sample 20 images for each object from the RGB multiview images for the image branch.

\paragraph{Scanobjectnn \cite{scanobjectnn}} is a dataset of real scans of 15 object classes. 10 2D views per each object are generated using \cite{pointclipv2}. We report results on 3 different versions of the dataset; Sc-obj - clean point cloud objects, Sc-obj withbg - Scans of objects with backgrounds, and Sc-obj hardest - Scans with backgrounds and additional random scaling and rotation augmentations.

\begin{table}[h!]
\centering
\setlength{\tabcolsep}{1mm}{
\resizebox{1\linewidth}{!}{
\begin{tabular}
{l|cc|cc|c}
\toprule
& \textbf{Train} & & \textbf{Test} & & \\
 & Images & Point clouds & Images & Point clouds & Categories \\
\midrule
Shapenet & 503316 & 41943 & 126204 & 10517 & 55 \\
ModelNet & 38196 & 3183 & 9600 & 800 & 40 \\
Redwood & 17270 & 314 & 4620 & 84 & 9 \\
co3d & 110536 & 5519 & 28192 & 1406 & 42 \\
Scanobjectnn & 23090 & 2309 & 5810 & 581 & 15 \\
\bottomrule
\end{tabular}
}}
\caption{Dataset details}
\label{tab:dataset details}
\end{table}

\paragraph{Qualitative comparison:} qualitative comparison of the point clouds and accompanying 2D views for the aforementioned datasets are shown in Figure \ref{fig:qualitative-dataset}

\subsection{Modelnet40 dataset splits:}
There are some common classes between our pre-train dataset, ShapeNet55, and ModelNet40. Evaluations on these common classes might introduce an unfair comparison of zeroshot performance. ULIP \cite{ULIP} authors introduced three different validation sets for evaluating models and baselines on ModelNet40.

All Set: Includes all the categories in ModelNet40 as shown
in Table \ref{tab:Modelnet40-All}.

Medium Set: Categories whose exact category names exist in pre-training dataset; Shapenet55 have been removed. The resulting categories in this set are shown in Table \ref{tab:Modelnet40-Medium}.

Hard Set:Both extract category names and their synonyms in the pre-training dataset have been removed. The final Hard Set is shown in Table \ref{tab:Modelnet40-Hard}.

\begin{table}[h]
\centering
\setlength{\tabcolsep}{1mm}{
\resizebox{0.9\linewidth}{!}{
\begin{tabular}
{ccccc}
\toprule
Airplane & Bathtub & Bed & Bench & Bookshelf \\
\midrule
Bottle & Bowl & Car & Chair & Cone \\
\midrule
Cup & Curtain & Desk & Door & Dresser \\
\midrule
Flower Pot & Glass Box & Guitar & Keyboard & Lamp \\
\midrule
Laptop & Mantel & Monitor & Night Stand & Person \\
\midrule
Piano & Plant & Radio & Range Hood & Sink \\
\midrule
Sofa & Stairs & Stool & Table & Tent \\
\midrule
Toilet & TV Stand & Vase & Wardrobe & Xbox \\
\bottomrule
\end{tabular}
}}
\caption{\small Modelnet40-All set}
\label{tab:Modelnet40-All}
\end{table}

\begin{table}[h]
\centering
\setlength{\tabcolsep}{1mm}{
\resizebox{0.9\linewidth}{!}{
\begin{tabular}
{ccccc}
\toprule
Cone & Cup & Curtain & Door & Dresser \\
\midrule
Glass Box & Mantel & Monitor & Night Stand & Person \\
\midrule
Plant & Radio & Range Hood & Sink & Stairs \\
\midrule
Stool & Tent & Toilet & TV Stand & Vase \\
\midrule
Wardrobe & Xbox & & & \\
\bottomrule
\end{tabular}
}}
\caption{\small Modelnet40-Medium set}
\label{tab:Modelnet40-Medium}
\end{table}

\begin{table}[h]
\centering
\setlength{\tabcolsep}{1mm}{
\resizebox{0.9\linewidth}{!}{
\begin{tabular}
{ccccc}
\toprule
Cone & Curtain & Door & Dresser & Glass Box \\
\midrule
Mantel & Night Stand & Person & Plant & Radio \\
\midrule
Range Hood & Sink & Stairs & Tent & Toilet \\
\midrule
TV Stand & Xbox & & & \\
\bottomrule
\end{tabular}
}}
\caption{\small Modelnet40-Hard set}
\label{tab:Modelnet40-Hard}
\end{table}

\section{Measuring entropy and biasedness of predictions} \label{appendix:training_trends_equations}
To calculate the entropy and biasedness of the predictions of our model, we use KL-Divergence to model the relative entropy from $U$ to $P$ where $U$ and $P$ are probability distributions. 
$$ D_{KL}( P||U) =\sum _{x\in \mathcal{X}} P( x)\log\left(\frac{P(x)}{U( x)}\right)$$
First we define a uniform probability distribution $u$, where $C$ is the number of categories.
$$u=\left\{\frac{1}{C}\right\}_{c=1}^{C}$$
$p$ is the logits produced by the model for a single input in the test set. Then, the entropy of the particular prediction can be calculated as:
$$Pred.\ Entropy\ =\sum _{c\in C} p_{c}\log\left(\frac{p_{c}}{u_{c}}\right) =\sum _{c\in C} p_{c}\log( Cp_{c})$$ and averaged over the test set to record the entropy of predictions as training progresses. 

To calculate the biasness of the model towards predicting certain classes, we define a vector $j$:
$$j=\frac{1}{S}\{a_{1} ,a_{2} ,a_{3} ...,a_{C}\}$$ where $S$ is the size of the test set, and $a_c$ is the number of samples predicted as category $c$ by the model.
Then, the prediction bias of the model in one test cycle is calculated as:
$$ Pred.\ Bias\ =\sum _{c\in C} j_{c}\log\left(\frac{j_{c}}{u_{c}}\right) =\sum _{c\in C} j_{c}\log( Cj_{c})$$

\section{Analysis of feature embeddings} \label{appendix:tsne_embeddings}
\begin{figure}[h]
\centering
    \includegraphics[width=0.9\columnwidth]{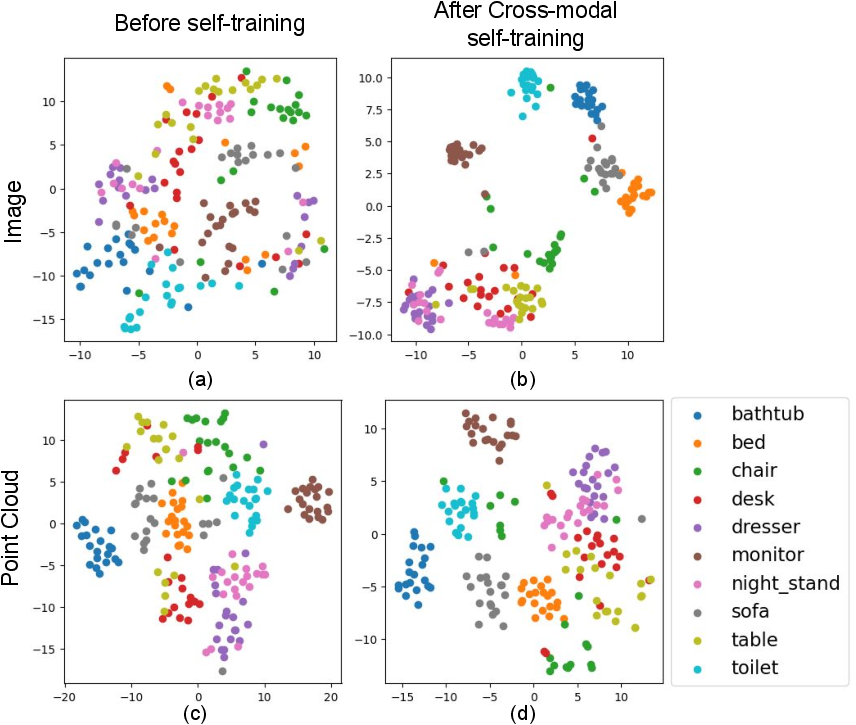}
  \caption{\small TSNE-feature embeddings for Modelnet10 before and after Cross-modal self-training.}
  \label{fig:TSNE-modelnet}
\end{figure}

\begin{figure}[h]
\centering
    \includegraphics[width=0.9\columnwidth]{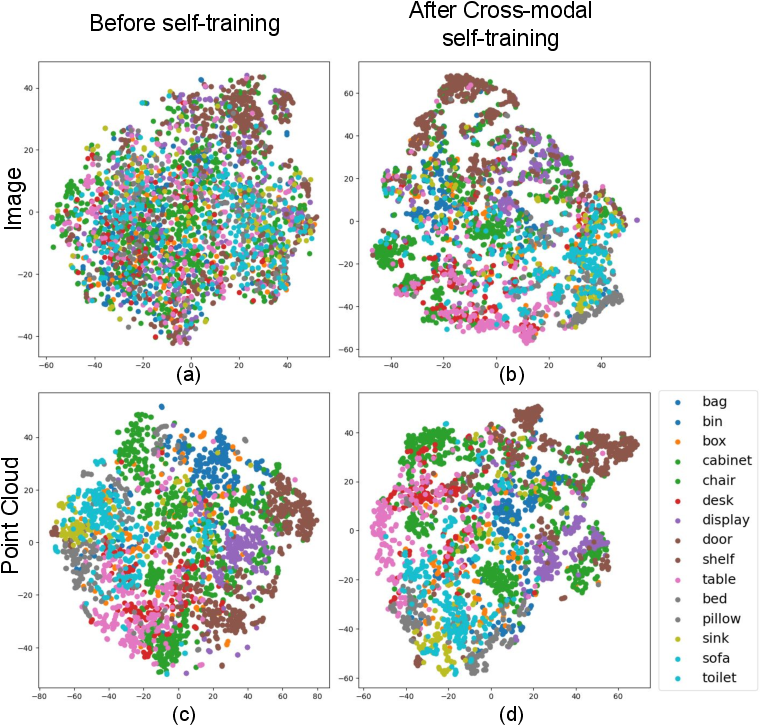}
  \caption{\small TSNE-feature embeddings for Scanobjectnn before and after Cross-modal self-training..}
  \label{fig:TSNE-scanobject}
\end{figure}

\paragraph{Quality of feature embeddings:} 
We visualize the TSNE embeddings of the 3D and 2D features extracted from respective modalities before and after cross-modal self-training for ModelNet10 in Figure \ref{fig:TSNE-modelnet} and Scanobjectnn \ref{fig:TSNE-scanobject}. 
In both cases, features from both image and point cloud modalities show improved class separability after self-training. 

\begin{figure*}[h]
\centering
{\includegraphics[width=0.85\textwidth]{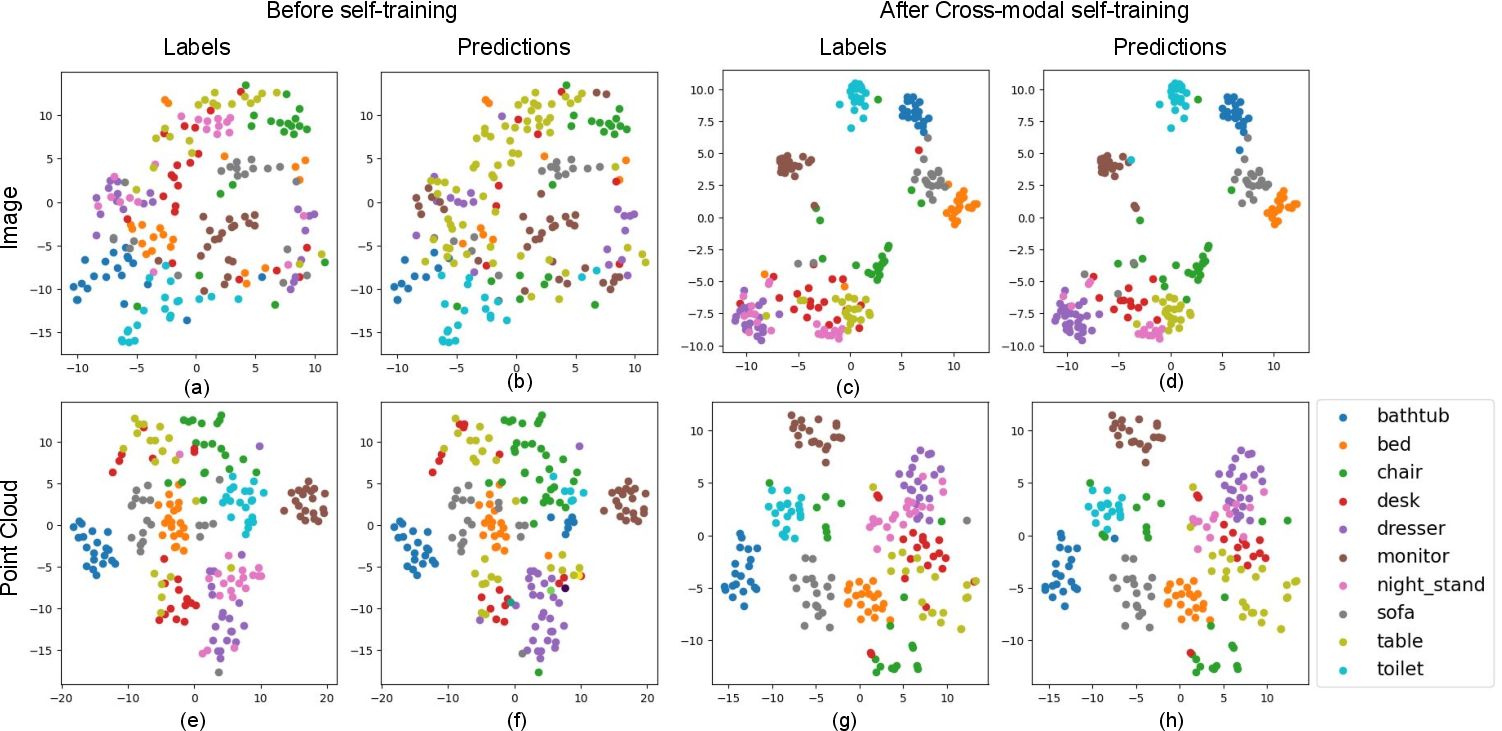}}
\caption{\small TSNE-feature embeddings for Modelnet10 before and after Cross-modal self-training with the original class labels and the predictions returned by our classifier. }  
\label{fig:TSNE-modelnet-withpred}
\end{figure*}

\begin{figure*}[h]
\centering
{\includegraphics[width=0.85\textwidth]{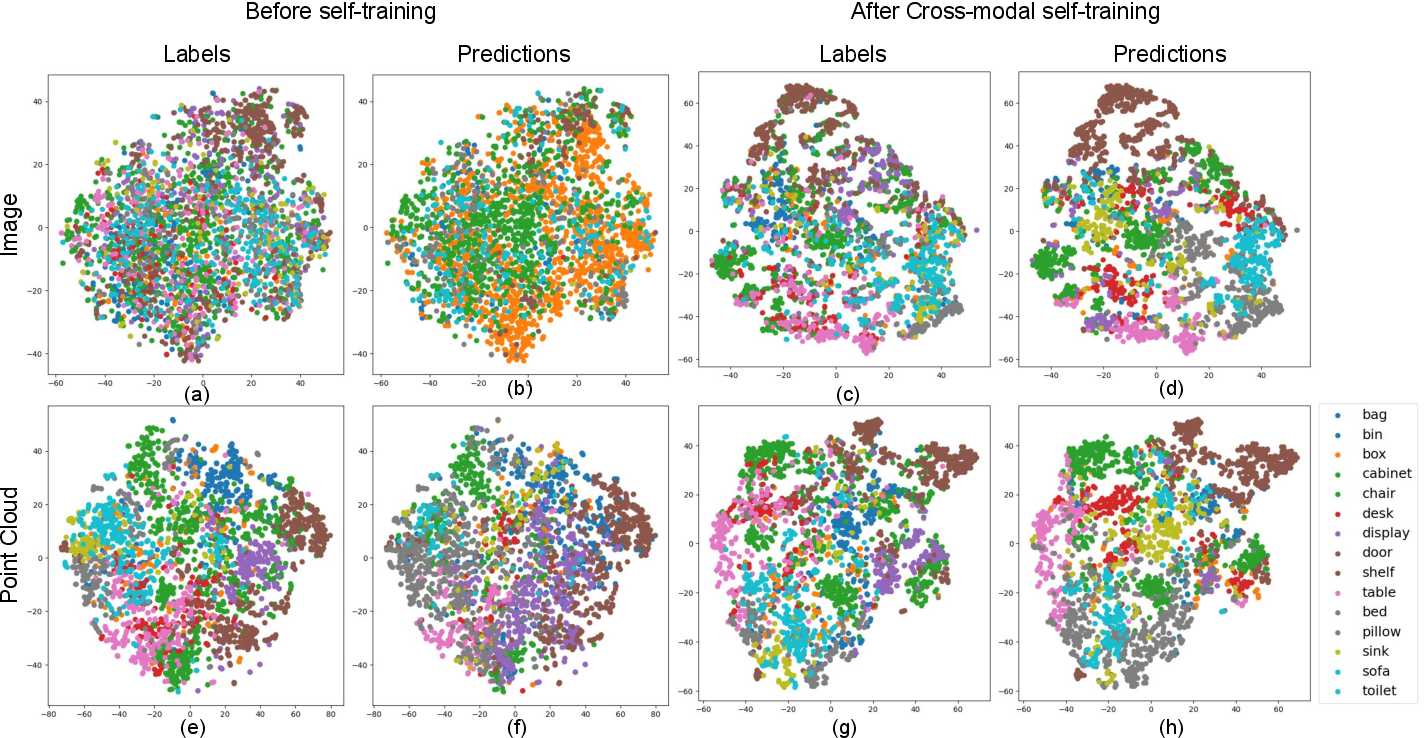}}
\caption{\small TSNE-feature embeddings for Scanobjectnn before and after Cross-modal self-training with the original class labels and the predictions returned by our classifier.}  
\label{fig:TSNE-scanobject-withpred}
\end{figure*}

Furthermore, in Figure \ref{fig:TSNE-modelnet-withpred} and Figure \ref{fig:TSNE-scanobject-withpred} we visualize the TSNE embeddings against the original class labels, as well as the predictions returned by our model. It is evident that the feature-level discrimination as well as class-level classification results are significantly improved over zeroshot by using our proposed method.

{
    \small
    \bibliographystyle{ieeenat_fullname}
    \bibliography{main}
}

\end{document}